\documentclass[conference]{IEEEtran}
\usepackage{times}

\usepackage[numbers]{natbib}
\usepackage{multicol}
\usepackage[bookmarks=true]{hyperref}
\usepackage{graphicx}
\usepackage{amsmath}
\usepackage{amsfonts}
\usepackage{amssymb}
\DeclareMathOperator*{\argmin}{arg\,min}
\usepackage[font=small]{caption}
\usepackage{multirow}
\usepackage{booktabs}
\usepackage{cite}
\usepackage[normalem]{ulem}
\usepackage{xcolor}
\usepackage{subfigure}

\begin{document}

\title{\LARGE \bf Early Failure Detection in Autonomous Surgical \\Soft-Tissue Manipulation via Uncertainty Quantification}

\author{Jordan Thompson, Ronald Koe, Anthony Le, Gabriella Goodman, Daniel S. Brown, and Alan Kuntz}

\input{intro-fig}
\maketitle
\thispagestyle{empty}
\pagestyle{empty}

\begin{abstract}

Autonomous surgical robots are a promising solution to the increasing demand for surgery amid a shortage of surgeons. Recent work has proposed learning-based approaches for the autonomous manipulation of soft tissue. However, due to variability in tissue geometries and stiffnesses, these methods do not always perform optimally, especially in out-of-distribution settings. We propose, develop, and test the first application of uncertainty quantification to learned surgical soft-tissue manipulation policies as an early identification system for task failures. We analyze two different methods of uncertainty quantification, deep ensembles and Monte Carlo dropout, and find that deep ensembles provide a stronger signal of future task success or failure. We validate our approach using the physical daVinci Research Kit (dVRK) surgical robot to perform physical soft-tissue manipulation. We show that we are able to successfully detect out-of-distribution states leading to task failure and request human intervention when necessary while still enabling autonomous manipulation when possible. Our learned tissue manipulation policy with uncertainty-based early failure detection achieves a zero-shot sim2real performance improvement of 47.5\% over the prior state of the art in learned soft-tissue manipulation. We also show that our method generalizes well to new types of tissue as well as to a bimanual soft-tissue manipulation task.

\end{abstract}

\section{Introduction}
	
Autonomous surgical robots have the potential to help solve the growing disparity between the population's need for surgery and the number of available surgeons~\citep{zhang2020physician,attanasio2021autonomy}. However, surgical robot learning and automation is particularly challenging due to the nuanced and risk-sensitive nature of the tasks, the partially observable and deformable nature of the environment, and the scarcity of available data. In particular, because autonomous surgical system failures can be detrimental to patient health, it is crucial that these autonomous systems take into account uncertainty so that the system can safely cede control to the surgeon before causing any harm. Thus, we aim to develop an autonomous surgical soft-tissue manipulation system that is capable of reasoning over predictive uncertainty to detect task failure early in the robot's operation (see Fig.~1). In doing so, our goal is to mitigate the risk of system failures while still offloading the work of soft-tissue manipulation to the robot, whenever it is safe to do so.

Our main goal is to enable uncertainty-based task failure prediction in the domain of soft-tissue manipulation. To this end, we build on recent advances in robot learning for deformable object manipulation. In particular, we focus on the recently proposed, state-of-the-art DeformerNet framework~\citep{thach2023deformernet} which uses large-scale self-supervised training in simulation to learn a closed-loop policy for manipulating deformable objects to desired goal geometries. While DeformerNet has been shown to perform well on in-distribution tissue, inter-human variability can lead to erroneous outputs, which can pose a safety risk. We propose to address this issue through quantifying the predictive uncertainty of DeformerNet.

One of the major obstacles when applying robot learning to surgical domains is the inability to do online learning in the real world. While we can collect demonstrations offline~\citep{pomerleau1991efficient,torabi2018behavioral,florence2022implicit,zhang2023discriminator} and can train policies in simulation~\citep{zhao2020sim,pashevich2019learning,kaspar2020sim2real,sharma2022learning}, offline imitation learning leads to compounding errors~\citep{dagger} and the sim2real gap is especially difficult to overcome for the types of deformable manipulation that are common in surgery~\citep{haiderbhai2022robust}. Enabling a surgical robot to preemptively detect task failures addresses these challenges by enabling a human to correct the robot during real-world execution, thus preventing failures while also enabling the robot to potentially use human interventions to continually improve its policy and world model. While there has been prior work on interactive imitation learning and human interventions~\citep{hg_dagger,safe_dagger, EIL,HITL,hoque2021thriftydagger,datta2023iifl}, to the best of our knowledge, we are the first to study robot-requested interventions on the kinds of complex deformable object manipulation tasks common in surgical robotics.

Our main contributions are as follows: (1) We develop and study the first real-world surgical robotic system capable of uncertainty quantification and early task failure detection during soft-tissue manipulation. (2) We extend and analyze two potential methods for quantifying predictive uncertainty during learned soft-tissue manipulation: deep ensembles and Monte Carlo dropout. We provide empirical evidence that ensembles provide more accurate uncertainty estimates and that using the slope of the ensemble variance rather than the raw variance more accurately discriminates between successful and unsuccessful tissue manipulation actions. (3) Across 40 real-world tissue manipulation trials, we show that our approach improves zero-shot sim2real performance of more than 47\% when compared with DeformerNet by itself, the prior state-of-the-art approach for deformable tissue manipulation.
Concretely, our method achieves a 95\% system success rate compared with DeformerNet achieving 47.5\% system success rate without our method.
Further, our learned thresholds for failure detection successfully generalize to out-of-distribution tissue and to new tasks unseen during training and calibration.

\section{Related Work}
\subsection{Soft Tissue Manipulation}

Much work has investigated automated tissue manipulation during surgery~\citep{attanasio2020autonomous}. Several data-driven approaches have been proposed to learn soft tissue manipulation~\citep{pore2021learning,shin2019autonomous,pedram2020toward,retana2022autonomous,thach2023deformernet}. These learning-based approaches are made possible by taking advantage of recent advancements in high-fidelity deformable object simulation~\citep{erleben2019non, huang2021defgraspsim, liang2018gpu}. Other work has shown success in using model-independent deformation estimation techniques~\citep{alambeigi2018toward}. Our method primarily draws on the recently proposed DeformerNet framework~\citep{thach2023deformernet}. DeformerNet takes a self-supervised learning approach to the problem of soft tissue manipulation. Given the current geometry of the object, the grasping point, and the desired geometry of the object, DeformerNet has been shown to have state-of-the-art performance for a variety of deformable object geometries and tasks~\citep{thach2023deformernet}. However, as we show in our experiments, DeformerNet sometimes fails due to out-of-distribution inputs or bad grasp points. By quantifying the uncertainty in DeformerNet's predictions, we enable informed early failure detection, taking an important step toward safe autonomy in soft-tissue surgical manipulation tasks. 

\subsection{Uncertainty Quantification}

Uncertainty quantification for deep neural networks has been studied in recent years as a way to aid in the safe design and implementation of deep learning systems~\citep{gawlikowski2023survey}. We study and compare the efficacy of ensembles and Monte Carlo dropout for uncertainty quantification for deformable robot manipulation tasks with a focus on soft-tissue manipulation. Ensembles are an approach by which multiple models are used in concert to make predictions and have been shown to improve overall model performance~\citep{lee2015m,fort2019deep}. Uncertainty is typically quantified using ensembles by measuring the variance in the ensemble's outputs~\citep{pearce2018high,lakshminarayanan2017simple}.

Monte Carlo dropout is another technique utilized for uncertainty estimation in deep neural networks~\citep{milanes2021monte,camarasa2020quantitative}. 
By extending dropout to the inference phase and making multiple inference passes through the network with different dropout instantiations, Monte Carlo dropout can be used to generate a distribution of outputs. Unlike ensembles, which require training multiple independent models, Monte Carlo dropout leverages the same trained model for inference, eliminating the need for repeated training~\citep{mae2021uncertainty}.  
 
By analyzing the distribution from an ensemble or Monte Carlo dropout, uncertainty metrics such as entropy or variance across predictions can be computed, providing insights into the model's confidence and reliability~\citep{sensoy2018evidential, mena2021survey}.

\subsection{Human Interventions for Robot Policies}
Offline learning from human demonstrations is a common approach for robot learning~\citep{pomerleau1991efficient, ho2016generative, argall2009survey, osa2018algorithmic, arora2018survey, ijspeert2013dynamical,paraschos2013probabilistic}; however, while learning from offline data has shown success in some domains~\citep{one-shot-visual-IL, rope-descriptors, ganapathi2020learning, bc_driving}, executing policies learned from offline data is known to lead to suboptimal behavior due to compounding errors~\citep{dagger, DART}, something that is unacceptable in surgical robotics. To remedy this problem, it is common practice to give control to a human supervisor who can provide an intervention. There are two strategies for deciding when to pass control from the autonomous agent to the human. In the first paradigm, the human decides when to intervene~\citep{hg_dagger, EIL, HITL}. However, this imposes a large burden on the supervisor since it requires a human to continuously monitor the robot and relies on the human accurately predicting what the robot will do next. In the second paradigm, the robot actively requests human interventions~\citep{ensemble_dagger,safe_dagger,hoque2021thriftydagger,datta2023iifl}. Similar to our approach, prior methods of this type typically use some form of novelty or uncertainty estimation when deciding to cede control to the human. However, prior work has focused mainly on simulated tasks with simulated human supervisors and simple control or manipulation tasks. By contrast, we study the efficacy of uncertainty quantification when deployed on a surgical dVRK robot performing complex, deformable tissue manipulation in the real world with access to real human interventions.  

\section{Problem Definition}
Given a learned surgical robot policy, $\pi_{\rm robot}$, for soft-tissue manipulation, we wish to develop an early identification system to determine whether the robot should execute its policy autonomously or request an intervention from a human supervisor. Following prior work on deformable tissue manipulation~\citep{thach2023deformernet}, we assume access to a partial-view point cloud of the current geometry of the soft tissue to be manipulated as well as a goal point cloud that defines how the soft-tissue should be manipulated.

The meta policy $\pi_{\rm meta}$ chooses whether to execute the learned robot policy or to hand over control to a human supervisor and can thus be represented as a binary classifier, $\pi_{\rm meta}: \mathcal{O} \rightarrow \{0,1\}$, where $\mathcal{O}$ is the observation space. A true positive (TP) is when the robot correctly requests an intervention and would have failed if it had operated autonomously, a false positive (FP) is when the robot wrongly asks for an intervention when it would have succeeded autonomously, a false negative (FN) is when the robot does not ask for an intervention and fails at the task, and a true negative (TN) is when the robot does not request an intervention and succeeds at the task autonomously.   

Because we are focused on surgical robotic applications, we are most concerned with failures. Thus, we want to minimize failures associated with false negatives. However, this is not the only source of potential failures---we also want to minimize false positives. False positives may also lead to failure because they unnecessarily distract a human expert and waste time that could be spent addressing other concerns (e.g., assisting a different patient or robot) and preventing other failures. Thus, given a cost of failure $c_f$, we want to find the following meta policy: 
\begin{equation}
\pi^*_{\rm meta} = \argmin_{\pi_{\rm meta}} \mathbb{E}[ c_f \cdot FN + c_f \cdot P(\rm failure | FP) \cdot FP],
\end{equation}
where $P(\rm failure|FP)$ is the probability of a failure occurring somewhere else due to the human supervisor being unnecessarily distracted or burdened.
Since $c_f > 0$, we can just minimize $FN + P(\rm failure | FP) \cdot FP$. We do this by learning the optimal classification threshold for $\pi_{\rm meta}$ using a calibration set~\citep{flach2016roc,koyejo2014consistent,pleiss2017fairness} to appropriately tradeoff between FN and FP. Note that the appropriate tradeoff will depend on the particular context and will be problem dependent.

\section{Methodology}

\subsection{DeformerNet}
For this work, we modify and endow the DeformerNet framework~\citep{thach2023deformernet} with uncertainty awareness and the ability to preemptively detect task failures conditioned on that uncertainty. DeformerNet is a deep learning based shape-servoing model for manipulating deformable objects. The model takes as input the current geometry $\mathrm{P}_c$ and goal geometry $\mathrm{P}_g$ of the deformable object as partial-view point clouds along with the manipulation point $\boldsymbol{m}$ on the object. The model then predicts an action $\hat{\mathrm{A}}$ as a homogeneous transformation matrix for the robot's end-effector,
$\hat{\mathrm{A}} = \begin{bmatrix}
    \hat{\mathrm{R}} & \hat{\boldsymbol{p}} \\
    \boldsymbol{0} & 1
\end{bmatrix}$,
where $\hat{\mathrm{R}}$ and $\hat{\boldsymbol{p}}$ are the predicted change in end-effector orientation and position respectively. 
DeformerNet is a closed loop control method that reasons over the deformable object's geometry and the goal shape's geometry leveraging encoders based on PointConv~\citep{wu2019pointconv}. DeformerNet is trained in a self-supervised fashion using large-scale simulated data. In simulation, given a manipulation point $\boldsymbol{m}$ on a deformable object with current geometry $\mathrm{P}_c$, a random homogeneous transformation matrix A is applied to the robot's end effector. The geometry of the deformable object is then sensed and is used as a self-supervised goal, $\mathrm{P}_g$. DeformerNet is trained to predict the homogeneous transformation that led to $\mathrm{P}_g$ over a data set consisting of instances of the following form: $(\mathrm{P}_c, \mathrm{P}_g, \boldsymbol{m}, \mathrm{A})$.

\begin{figure*}
    \centering
    \includegraphics[width=0.245\textwidth]{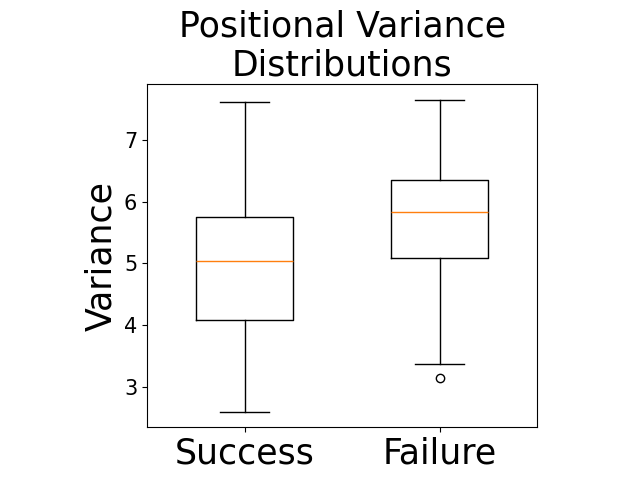}
    \includegraphics[width=0.245\textwidth]{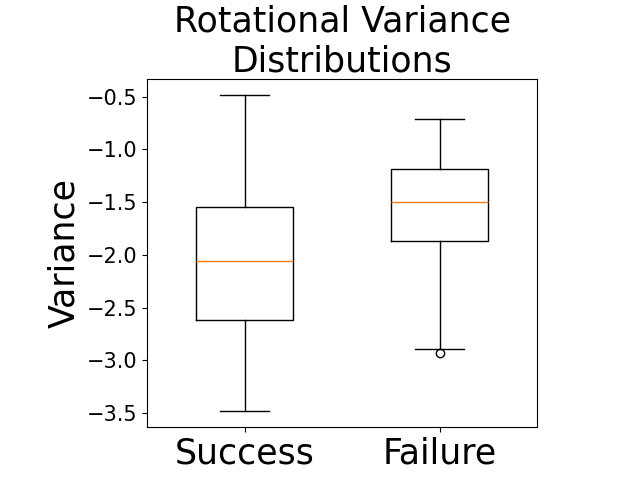}
    \includegraphics[width=0.245\textwidth]{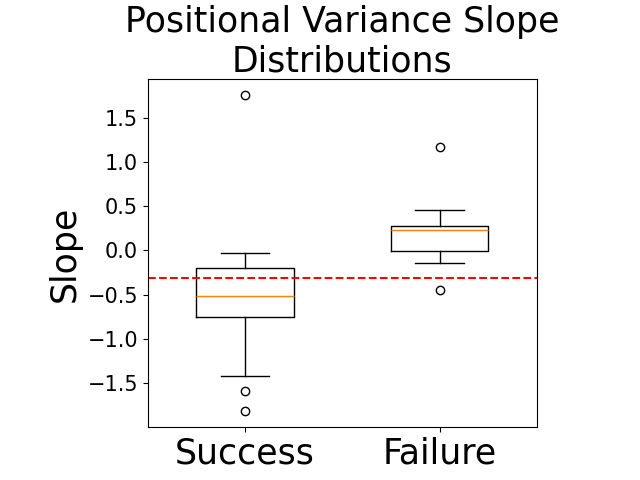}
    \includegraphics[width=0.245\textwidth]{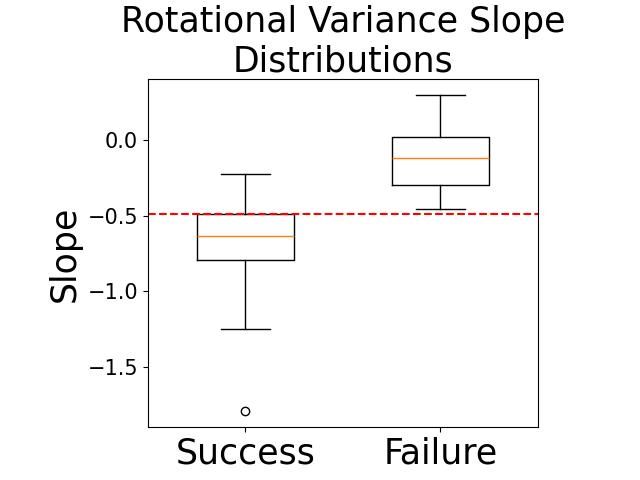}
    \caption{Distributions of raw variance values (computed via Equations~\ref{eq:pos_var} and~\ref{eq:rot_var}) and initial slopes (computed via Equation~\ref{eq:uncertainty} at $t=1$) for both position and rotation across 30 trials at inference time on the dVRK surgical robot. We find that the distributions of raw variance values for position and rotation across successful and failure trials only have Kullback-Leibler (KL) divergences of 0.357 and 0.507 respectively. However, the initial slopes of the variance values have KL-divergence values of 3.107 and 4.078 respectively. We set a threshold value on the slope of the initial positional and rotational variance values at -0.310 and -0.487 respectively as indicated by the red lines in the graphs.}
    \label{fig:var_distrib}
    \vspace{-1em}
\end{figure*}

\subsection{Uncertainty quantification}

We aim to augment the DeformerNet model with uncertainty quantification to enable early task failure detection. To this end, we propose the use of a deep ensemble consisting of multiple DeformerNet models. Each model predicts a change in position and orientation of the robot's end effector that will deform the object being manipulated toward the goal shape.
At inference time, the positional component $\hat{\boldsymbol{p}}$ of the deep ensemble's prediction is computed by averaging the positional component of the individual models, 
\begin{equation}
    \hat{\boldsymbol{p}} = \frac{1}{N}\sum_{i=1}^N{\hat{\boldsymbol{p}}_i} \;,
    \label{eq:pos_mean}
\end{equation}
where $N$ is the number of models in the ensemble, and $\hat{\boldsymbol{p}}_i$ is the positional output of model $i$.

At inference time, the orientation component of the deep ensemble's output is found by first computing the arithmetic mean $\mathrm{S}$ of the set of rotation matrices produced by the individual models of the ensemble,
$\mathrm{S} = \frac{1}{N}\sum_{i=1}^N{\hat{\mathrm{R}}_i}$.
where $N$ is the number of models, and $\hat{\mathrm{R}}_i$ is the rotation matrix output by model $i$. Matrix $\mathrm{S}$, however, will almost certainly not satisfy the constraints required to form a valid rotation matrix. To obtain a valid rotation matrix, we take the singular value decomposition, $\mathrm{S} = \mathrm{U}\mathrm{D}\mathrm{V}'$ and
multiply $\mathrm{U}$ and $\mathrm{V}'$ to obtain a rotation matrix that minimizes the Euclidean norm to $\mathrm{S}$ and is therefore the rotation matrix that minimizes the average geodesic distance to each component rotation matrix~\citep{sarabandi2020closed}:
\begin{equation}
    \hat{\mathrm{R}} = \mathrm{U}\mathrm{V}' \;.
    \label{eq:rot_mean}
\end{equation}

We compute the ensemble's predictive variance for both position and rotation individually. Positional variance is the average squared distance to the deep ensemble's positional prediction ($\hat{\boldsymbol{p}}$ in Equation~\eqref{eq:pos_mean}) for each component network: 
\begin{equation}
\mathrm{Var}(\hat{\boldsymbol{p}}) = \frac{1}{N}\sum_{i=1}^N{\Vert\hat{\boldsymbol{p}}_i - \hat{\boldsymbol{p}}\Vert^2} \; .
\label{eq:pos_var}
\end{equation}

For rotational variance, we measure the average geodesic distance to the ensemble's predicted rotation matrix ($\hat{\mathrm{R}}$ in Equation~\eqref{eq:rot_mean}):  
\begin{equation}
\mathrm{Var}(\hat{\mathrm{R}}) = \frac{1}{N}\sum_{i=1}^N{\arccos\bigl(\frac{\mathrm{Tr}(\hat{\mathrm{R}}_i\hat{\mathrm{R}}) - 1}{2}\bigr)} \; .
\label{eq:rot_var}
\end{equation}

We define the predictive uncertainty of DeformerNet at time-step $t$ as a vector containing the change in both positional and rotational variances from equations~\eqref{eq:pos_var} and~\eqref{eq:rot_var}.
\begin{equation}
\begin{aligned}
\boldsymbol{u}(t) = [&\mathrm{Var}(\hat{\boldsymbol{p}}_t) - \mathrm{Var}(\hat{\boldsymbol{p}}_{t-1}), \\
                    &\mathrm{Var}(\hat{\mathrm{R}}_t) - \mathrm{Var}(\hat{\mathrm{R}}_{t-1})]
\label{eq:uncertainty}
\end{aligned}
\end{equation}

While standard approaches to uncertainty quantification use raw variance values as their uncertainty metric~\citep{hoque2021thriftydagger,gawlikowski2023survey,lakshminarayanan2017simple}, we find in our experiments that these raw variance values are not good indicators for downstream task performance. However, when we measure the slopes of these variance values over subsequent time steps, we find a significantly stronger signal of downstream task performance.

\subsection{Monte Carlo Dropout}

We also investigate uncertainty quantification using Monte Carlo dropout. Using stochastic dropout at inference time allows us to sample from a distribution of potential changes in end-effector pose. We then use equations~\eqref{eq:pos_mean} --~\eqref{eq:uncertainty} to get the model predictions and variances.

\begin{table*}   
  \centering
  \small
  \begin{tabular}{lrrrrrrrrrrrr}
    \toprule
    & \multicolumn{12}{c}{Intervention Requests} \\
    & \multicolumn{6}{c}{Chicken Tissue} & \multicolumn{6}{c}{Bovine Tissue} \\
     &\multicolumn{2}{c}{Positional }  &\multicolumn{2}{c}{Rotational } &\multicolumn{2}{c}{Both }      &\multicolumn{2}{c}{Positional }  &\multicolumn{2}{c}{Rotational } &\multicolumn{2}{c}{Both }\\ 
    \cmidrule(r){2-7} \cmidrule(r){8-13} 
    & Yes & No & Yes & No & Yes & No & Yes & No & Yes & No & Yes & No \\
    \midrule
    Intervention Needed  & \textbf{10} & 1 & \textbf{9} & 2 & \textbf{10} & 1 & \textbf{9} & 1 & \textbf{10} & 0 & \textbf{10} & 0 \\
 Intervention Not Needed & 3 & \textbf{6} & 5 & \textbf{4} & 6 & \textbf{3} & 1 & \textbf{9} & 3 & \textbf{7} & 3 & \textbf{7} \\ 
    \bottomrule
  \end{tabular}
  \vspace{2ex}
  \caption{Results from using variance slope thresholds for requests for human intervention on ex vivo chicken muscle tissue and ex vivo bovine tissue across 40 trials. Placing a threshold on the change in positional variance gives an accuracy of 85\% across ex vivo chicken and bovine tissue with a false positive rate of 21.1\% and a false negative rate of 9.5\%. Using rotational variance, we achieve an accuracy of 75\% with a false positive rate of 42.1\% and a false negative rate of 9.5\%. Simultaneously thresholding position and rotation yields an accuracy of 75\% with a false positive rate of 47.4\% and a false negative rate of 4.8\%.}
\label{tbl:chicken_beef_results}
\vspace{-1em}
\end{table*}

\section{Physical Experimental Setup}
\label{sec:phys_setup}

We train a deep ensemble DeformerNet consisting of five models. We use the source code and training and testing datasets provided by Thach et al.~\citep{thach2023deformernet}, consisting of 11,566 training and 1,285 test examples of manipulations of a deformable box object. Following prior work~\citep{lee2015m,lakshminarayanan2017simple,fort2019deep}, each ensemble component model was trained using the same training dataset with different random weight initializations. 
% Using the same training and testing datasets as described for the deep ensemble, 
We train a Monte Carlo dropout version of the DeformerNet model with dropout added to the last layer at varying levels of dropout percentage: 25\%, 50\%, and 75\%. We collect 100 samples from the model to construct the predictive distributions.

The deep ensemble and the Monte Carlo dropout models are implemented using a zero-shot sim2real framework. Trained entirely on a simulated box-shaped deformable object, the system is tasked with manipulating both ex vivo chicken and bovine tissue of varying geometries.
To generate goal shapes for evaluation purposes only, using ex vivo tissue we teleoperated the robot to manually manipulate the tissue to a desired goal geometry. 
We then reset the system and task the models with manipulating the tissue to the same desired geometry (with no knowledge of how the shape was generated). 
We use an Intel Realsense D405 camera for tracking point cloud representations of the tissue geometry both in goal generation for evaluation and during method execution. 
We track the model variance as it manipulates the tissue toward the goal shape and measure task success as whether the model converges to the desired geometry. We define a method termination criteria as when $\lVert\hat{\boldsymbol{p}}\rVert < 0.001$. At this point, the method has effectively stopped moving the robot's end effector. We then compare the resulting tissue geometry with the desired geometry to test for task success. Figure~1 shows an image of the experimental setup. We use the patient-side dVRK manipulator(s) to manipulate the tissue both autonomously and by teleoperation.

\section{Physical Soft-Tissue Manipulation Results}
\label{sec:phys_results}

\subsection{DeformerNet Performance Validation}

\paragraph{In-Distribution Performance}
In prior work, DeformerNet has been shown to be capable of achieving 100\% success rates on in-distribution cases~\citep{thach2023deformernet}. To validate that our ensemble is capable of achieving similar performance, we performed 20 in-distribution trials on ex vivo chicken muscle tissue with optimal grasping points and goal geometries that only required local control. Our ensemble successfully manipulated the tissue in all 20 in-distribution trials.
We also performed this same experiment across 10 in-distribution cases using the Monte Carlo dropout model. We find that the Monte Carlo dropout version of DeformerNet fails to complete the task in all cases. Thus, applying Monte Carlo dropout at inference time severely hinders the performance of DeformerNet. As a result, the following experiments are performed solely using the deep ensemble.

\paragraph{Out-of-Distribution Performance}
While DeformerNet has 100\% task success on in-distribution cases, out-of-distribution cases frequently result in task failure. We performed 15 trials spanning 3 different out-of-distribution cases: sub-optimal grasping points, non-local control, and out-of-distribution tissue geometries. Sub-optimal grasping points can occur when the robot grasps the tissue at a location that does not optimally facilitate the manipulation of the tissue to the goal. Non-local control cases occur when the requested goal geometry requires more than one type of manipulation in sequence, i.e. folding the tissue which requires first lifting an edge then subsequently placing it back down elsewhere. Out-of-distribution tissue geometries occur when the geometry of the tissue being manipulated does not align with the tissue geometries seen during model training. Our ensemble only succeeded in 1 out-of-distribution geometry case while failing in the remaining 14 cases. Due to the high variability and lack of available data on surgical soft-tissue manipulation, these out-of-distribution cases will inevitably occur in practice. Thus, there is a great need to detect out-of-distribution cases prior to the manipulation of the tissue.

\subsection{Uncertainty Quantification}

We collected 30 trials containing both in-distribution and out-of-distribution cases of DeformerNet on ex vivo chicken muscle tissue and measured the predictive variances across the trajectories using Equations~\ref{eq:pos_var} and~\ref{eq:rot_var}. Figure~\ref{fig:var_distrib} shows the distributions of these raw variance values for successful and unsuccessful trials. We find that raw variance values are not good indicators of task success, as there is no clear distinction between the successful and unsuccessful trials.

Using our uncertainty definition from Equation~\ref{eq:uncertainty}, however, we find that the slopes in the variance values at $t=1$ provide a strong predictive signal of task success. We use Kullback-Leibler (KL) divergence as a quantitative metric of distribution difference. In the case of raw variance values, we observe KL divergences of 0.357 and 0.507 for position and rotation respectively. However, the slopes in variance at $t=1$ give KL divergences of 3.107 and 4.078 for position and rotation respectively. These results demonstrate that in successful cases, the positional and rotational variances tend to decrease moreso than in failure cases at the initial time step.

\begin{figure}
    \centering
    \includegraphics[width=\columnwidth]{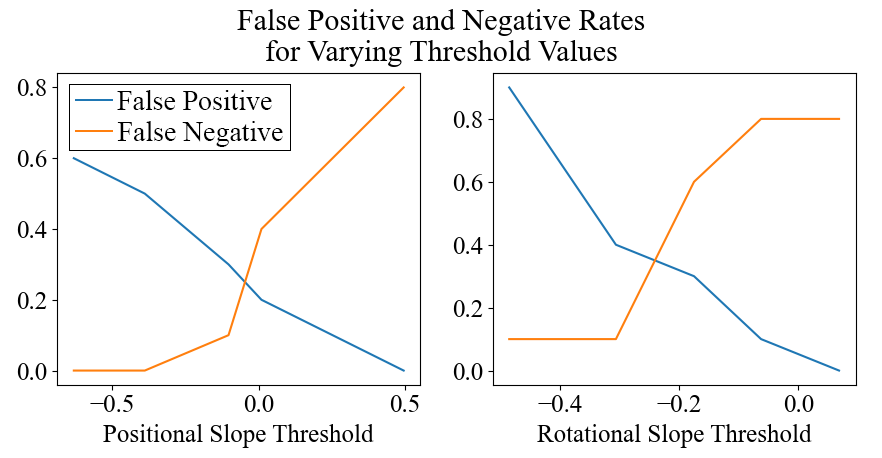}
    \caption{False positive and negative rates on 20 ex vivo chicken tissue trials for various positional and rotational slope thresholds. As the thresholds are raised, the false positive rates decrease while the false negative rates increase.}
    \label{fig:thresh_exp}
    \vspace{-1em}
\end{figure}

\paragraph{Calibration of the Meta Policy}
To detect task failures, we set a threshold on the slope of the positional and rotational variances at time-step 1 in the trajectories ($\boldsymbol{u}(1)$ from Eq.~\ref{eq:uncertainty}) by calibrating to desired false positive and negative rates on the above 30 trials (false positives are detecting task failure in successful cases, false negatives are failing to detect task failure when necessary). False positive and negative rates for different threshold values on 20 test trials can be seen in Fig.~\ref{fig:thresh_exp}.

\paragraph{Evaluation on Chicken Tissue} Placing thresholds at -0.310 and -0.487 for the positional and rotational components of $\boldsymbol{u}(1)$ respectively, we ran 20 trials on ex vivo chicken muscle tissue. The system detects task failure when the specified threshold is violated. Once the system detects task failure, the human teleoperates the remainder of the task. The results of these trials are summarized in Table~\ref{tbl:chicken_beef_results}. When only using the positional threshold, we successfully classified the success of the task in 16 out of the 20 trials. In only 1 out of the 20 trials we did not detect a task failure. When only using the rotational threshold, we successfully classified task success in 13 out of 20 trials while not detecting task failure in 2 of the trials. When using a threshold on both position and rotation simultaneously, we successfully classify task success in 13 trials and do not detect task failure in 1 trial.

\paragraph{Generalization to Bovine Tissue} To test the generalization of our method to new tissue types, we ran 20 trials on ex vivo bovine tissue with no re-calibration of the chosen thresholds. The results for the bovine tissue are summarized in Table~\ref{tbl:chicken_beef_results}. We successfully classify task completion in 18 and 17 of the trials for positional and rotational intervention respectively while only failing to detect a task failure in 1 of the positional intervention trials. When using both thresholds, we successfully classify task completion in 17 trials and never fail to detect a task failure.

\begin{figure}
    \centering
    \includegraphics[width=0.9\columnwidth]{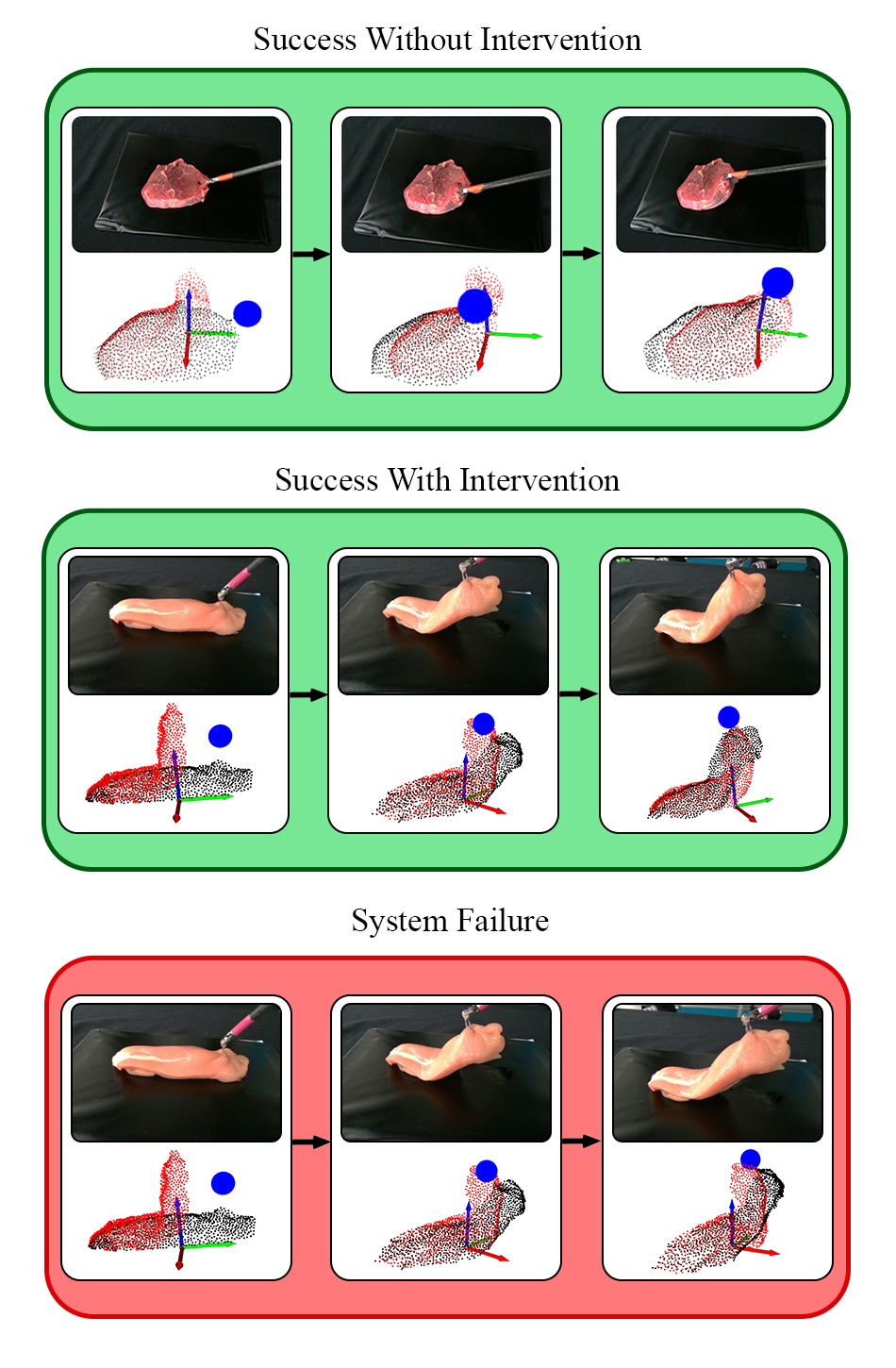}
    \caption{Example soft-tissue manipulations with and without requests for human intervention. The images show the view of the RealSense camera at subsequent steps of the trajectory. The point clouds show the corresponding current (black) and goal (red) geometries of the tissue while the grasp point on the tissue is shown as a blue dot. (Top) A successful manipulation where human intervention was not requested. (Middle) A successful manipulation where human intervention was requested. (Bottom) A failed manipulation where the trajectory from the middle panel is allowed to complete autonomously instead of having a human intervene.}
    \label{fig:ex_traj}
    \vspace{0em}
\end{figure}

Across all 40 physical trials, the baseline DeformerNet failed to reach the desired geometry in 52.5\% of trials. With the introduction of uncertainty thresholds, we successfully reduced the system failure rate to 5\% while using either positional or rotational thresholds constituting a 47.5\% improvement over base DeformerNet while still allowing autonomous manipulation in approximately 78.9\% of the trials for position, 57.9\% of the trials for rotation, and 52.6\% of the trials for both. Figure~\ref{fig:ex_traj} shows examples of both successful and failed manipulations with and without human intervention. We see that we are successfully able to enable autonomous tissue manipulation when appropriate while preventing system failures by requesting human intervention when necessary.

\begin{figure}
    \centering
    \includegraphics[width=0.9\columnwidth]{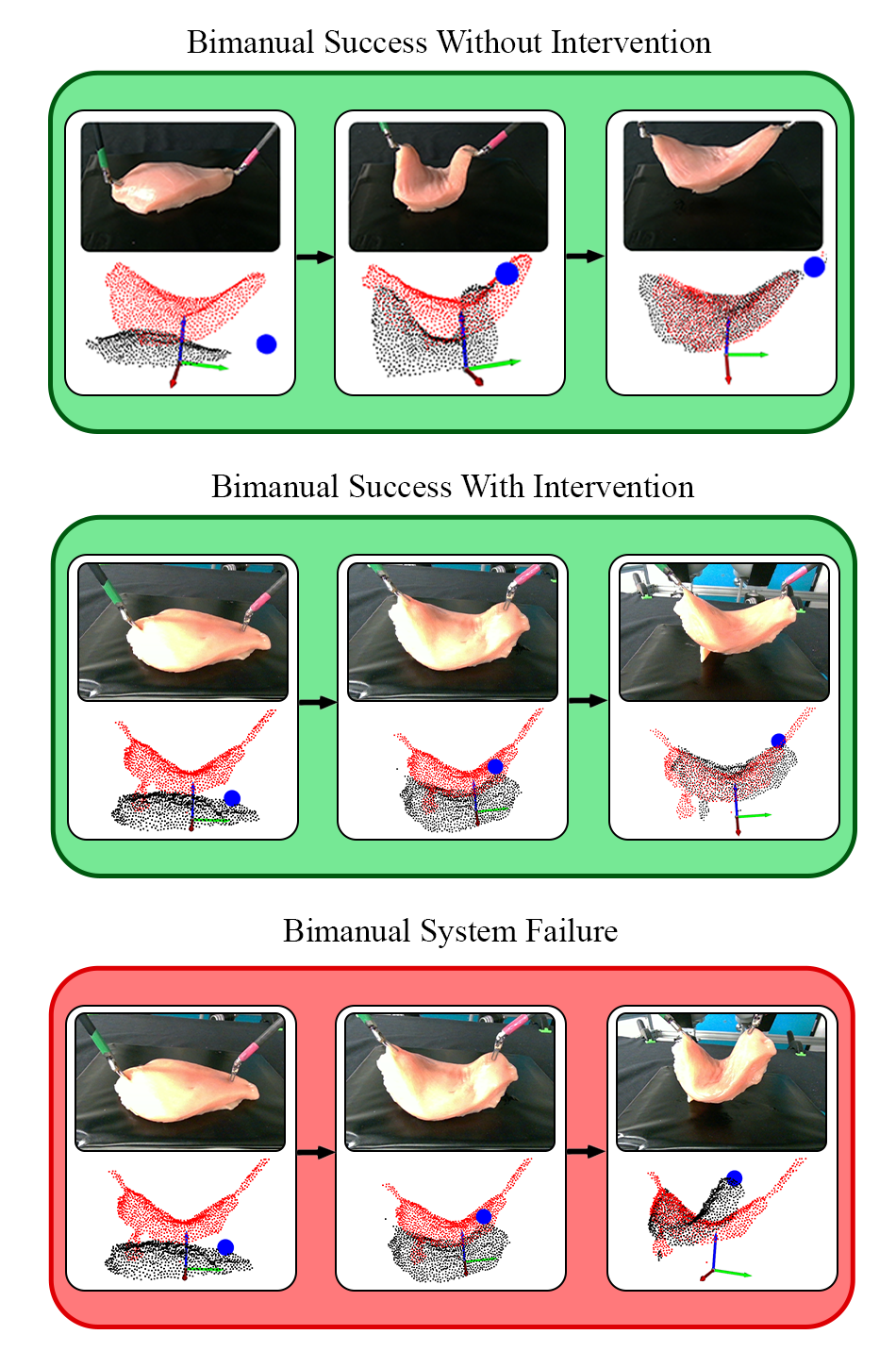}
    \caption{Example bimanual soft-tissue manipulations. The images show the view of the RealSense camera at subsequent steps of the trajectory. The point clouds show the corresponding current (black) and goal (red) geometries of the tissue while the grasp point on the tissue is shown as a blue dot. The left patient-side manipulator of the dVRK is always teleoperated by a human while the right patient-side manipulator is autonomously controlled by DeformerNet; however, the right patient-side manipulator can request a human to teleoperate when needed. (Top) A successful manipulation where human intervention was not requested. (Middle) A successful manipulation where human intervention was requested. (Bottom) A failed manipulation where the trajectory from the middle panel was instead allowed to complete the task autonomously despite requesting human intervention.}
    \label{fig:bimanual}
    \vspace{-1em}
\end{figure}

\paragraph{Task Generalization} We also tested our method's ability to generalize to new tasks by applying our thresholds to a bimanual shape-servoing task. In this task, the left patient-side manipulator of the dVRK is always teleoperated by the human while the right patient-side manipulator is controlled via our DeformerNet ensemble. The human operator is given control of the automated arm when an uncertainty threshold is violated. Similarly to the single-arm task, example bimanual soft-tissue manipulations can be seen in Fig.~\ref{fig:bimanual}. Across 10 trials with positional interventions, we achieved a 100\% system success rate with a false positive rate of 66.6\% and a false negative rate of 0\% (7 requested interventions while only 1 intervention was necessary). Using rotational interventions, we again achieved a 100\% system success rate with 0\% false positive and negative rates (1 requested intervention in a case where it was necessary). These results show that while our rotational threshold was transferable to this new task with no necessary recalibration, our positional threshold is overly conservative in this new task and would need to be recalibrated to achieve similar performance to the single-arm task.

\section{Conclusion}

Robot learning for the autonomous completion of surgical tasks is a particularly challenging domain due to data scarcity, out-of-distribution observations, and safety concerns. To address this challenge, we introduce a novel framework for the safe implementation of an autonomous soft-tissue manipulation system for uncertainty quantification and early failure detection. We apply our method on the dVRK surgical robot system for a soft-tissue manipulation task using both in- and out-of-distribution ex-vivo tissue types and demonstrate with high accuracy the ability to request interventions in cases that would have otherwise failed.
In the future we plan to study bi-directional human-robot hand-offs during soft tissue manipulation as well as study when humans feel confident ceding control back to the robot and how to visualize model uncertainty to enhance interpretability and trust.

\section*{Acknowledgments}
Research reported in this publication was supported by the Advanced Research Projects Agency for Health (ARPA-H) under Award Number D24AC00415-00. The ARPA-H award provided 100\% of the total costs with an award total of up to \$11,935,038. The content is solely the responsibility of the authors and does not necessarily represent the official views of ARPA-H.

\bibliographystyle{plainnat}
\bibliography{root}

\end{document}